\documentclass[10pt,twocolumn,letterpaper]{article}

\usepackage[pagenumbers]{cvpr} 
\makeatletter
\@namedef{ver@everyshi.sty}{}
\makeatother
\usepackage{tikz}
\usepackage{graphicx}
\usepackage{amsmath}
\usepackage{amssymb}
\usepackage{booktabs}
\usepackage{color}
\usepackage{mathrsfs}  
\usepackage{todonotes}
\usepackage{algorithmic}
\usepackage[linesnumbered,ruled,vlined]{algorithm2e}
\usepackage{multirow}
\usepackage{makecell}
\usepackage{pifont}
\usepackage{booktabs}
\usepackage{graphics}
\usepackage{pifont}
\usepackage{caption}
\usepackage{subcaption}
\definecolor{colivegreen}{rgb}{0,0.6,0}
\SetKwInput{KwInput}{Input}                
\SetKwInput{KwOutput}{Output}              

\SetCommentSty{mycommfont}

\newtheorem{assumption}{Assumption}

\usepackage[pagebackref,breaklinks,colorlinks]{hyperref}

\usepackage[capitalize]{cleveref}
\crefname{section}{Sec.}{Secs.}
\Crefname{section}{Section}{Sections}
\Crefname{table}{Table}{Tables}
\crefname{table}{Tab.}{Tabs.}

\def\cvprPaperID{5428} 
\def\confName{CVPR}
\def\confYear{2023}

\begin{document}

\title{Active Finetuning: Exploiting Annotation Budget \\ in the Pretraining-Finetuning Paradigm}

\author{Yichen Xie$^{1}$, Han Lu$^{2}$, Junchi Yan$^{2}$, Xiaokang Yang$^{2}$, Masayoshi Tomizuka$^{1}$, Wei Zhan$^{1}$\\ 
$^{1}$ University of California, Berkeley $^{2}$ Shanghai Jiao Tong University\\ 
{\tt\small \{yichen\_xie,tomizuka,wzhan\}@berkeley.edu,\{sjtu\_luhan,yanjunchi,xkyang\}@sjtu.edu.cn}
}

\maketitle

\begin{abstract}
Given the large-scale data and the high annotation cost, pretraining-finetuning becomes a popular paradigm in multiple computer vision tasks. Previous research has covered both the unsupervised pretraining and supervised finetuning in this paradigm, while little attention is paid to exploiting the annotation budget for finetuning. To fill in this gap, we formally define this new active finetuning task focusing on the selection of samples for annotation in the pretraining-finetuning paradigm. We propose a novel method called ActiveFT for active finetuning task to select a subset of data distributing similarly with the entire unlabeled pool and maintaining enough diversity by optimizing a parametric model in the continuous space. We prove that the Earth Mover's distance between the distributions of the selected subset and the entire data pool is also reduced in this process. Extensive experiments show the leading performance and high efficiency of ActiveFT superior to baselines on both image classification and semantic segmentation. Our code is released at \href{https://github.com/yichen928/ActiveFT}{https://github.com/yichen928/ActiveFT}.
\end{abstract}


\section{Introduction}
Recent success of deep learning heavily relies on abundant training data. However, the annotation of large-scale datasets often requires intensive human labor. This dilemma inspires a popular \textit{pretraining-finetuning paradigm} where models are pretrained on a large amount of data in an unsupervised manner and finetuned on a small labeled subset. 

\begin{figure}[tb!]
    \centering
    \includegraphics[width=\linewidth]{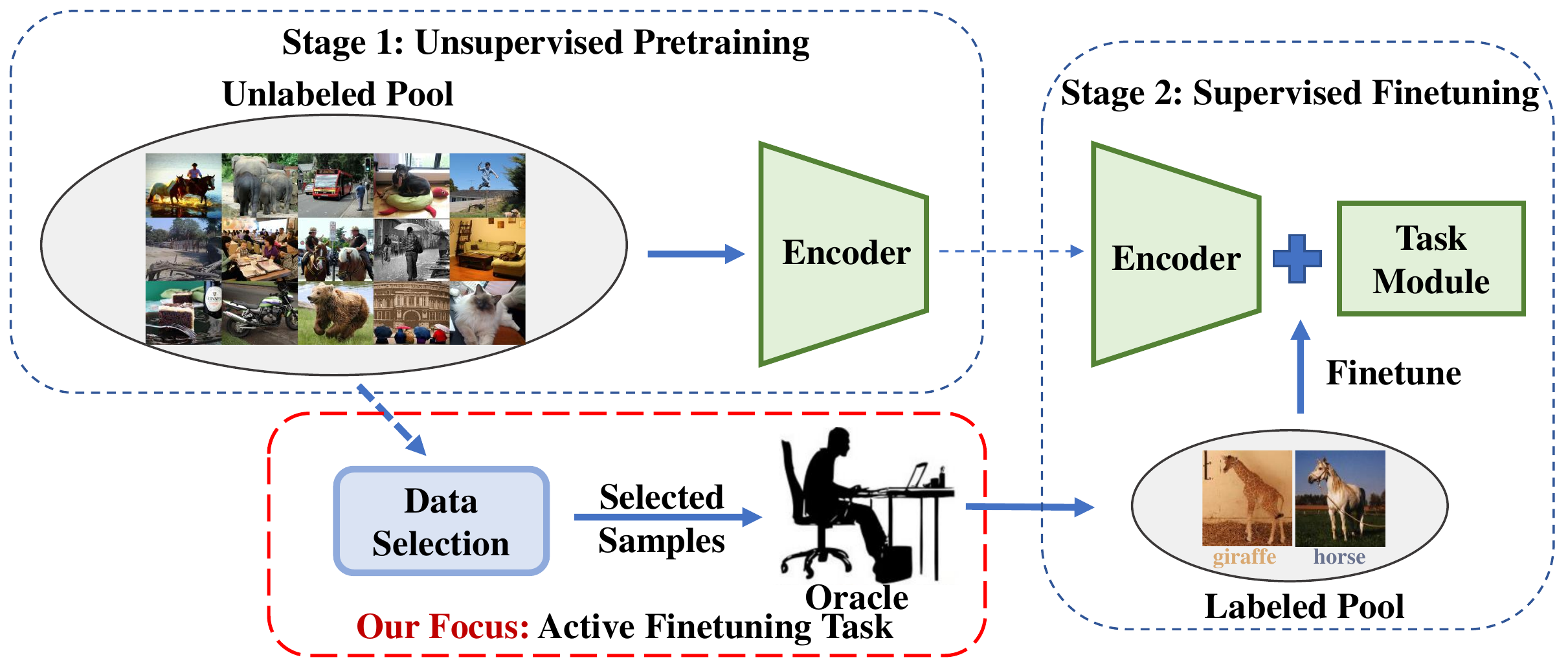}
    \caption{\textbf{Pretraining-Finetuning Paradigm:} We focus on the selection strategy of a small subset from a large unlabeled data pool for annotation, named as active finetuning task, which is under-explored for a long time.}
    \label{fig:paradigm}
\end{figure}

Existing literature pays significant attention to both the unsupervised pretraining \cite{chen2020simple,grill2020bootstrap,he2021masked,huang2021spatio} and supervised finetuning \cite{liu2021pre}. In spite of their notable contributions, these researches build upon an unrealistic assumption that \textit{we already know which samples should be labeled}. As shown in Fig.~\ref{fig:paradigm}, given a large unlabeled data pool, it is necessary to pick up the most useful samples to exploit the limited annotation budget. In most cases, this selected subset only counts a small portion (\textit{e.g.} $<$10\%) of this large unlabeled pool. Despite the long-standing under-exploration, the selection strategy is still crucial since it may significantly affect the final results.

Active learning algorithms \cite{ren2021survey,sener2017active,sinha2019variational} seem to be a potential solution, which aims to select the most suitable samples for annotation when models are trained from scratch. However, their failures in this pretraining-finetuning paradigm are revealed in both \cite{bengar2021reducing} and our experiments (Sec.~\ref{sec:classification}). A possible explanation comes from the \textit{batch-selection strategy} of most current active learning methods. Starting from a random initial set, this strategy repeats the model training and data selection processes multiple times until the annotation budget runs out.
Despite their success in from-scratch training, it does not fit this pretraining-finetuning paradigm well due to the typically low annotation budget, where too few samples in each batch lead to harmful bias inside the selection process.  

To fill in this gap in the pretraining-finetuning paradigm, we formulate a new task called \textit{active finetuning}, concentrating on the sample selection for supervised finetuning. In this paper, a novel method, \textbf{ActiveFT}, is proposed to deal with this task. Starting from purely unlabeled data, ActiveFT fetches a proper data subset for supervised finetuning in a negligible time. Without any redundant heuristics, we directly bring close the distributions between the selected subset and the entire unlabeled pool while ensuring the diversity of the selected subset. This goal is achieved by continuous optimization in the high-dimensional feature space, which is mapped with the pretrained model. 

We design a parametric model $p_{\theta_{\mathcal{S}}}$ to estimate the distribution of the selected subset. Its parameter $\theta_{\mathcal{S}}$ is exactly the high-dimensional features of those selected samples. We optimize this model via gradient descent by minimizing our designed loss function. Unlike traditional active learning algorithms, our method can select all the samples from scratch in a single-pass without iterative batch-selections. We also mathematically show that the optimization in the continuous space can exactly reduce the earth mover's distance (EMD) \cite{rubner2000earth,rubner1998metric} between the entire pool and selected subset in the discrete data sample space.

Extensive experiments are conducted to evaluate our method in the pretraining-finetuning paradigm. After pretraining the model on ImageNet-1k \cite{russakovsky2015imagenet}, we select subsets of data from CIFAR-10, CIFAR-100 \cite{krizhevsky2009learning}, and ImageNet-1k \cite{russakovsky2015imagenet} for image classification, as well as ADE20k \cite{zhou2017scene} for semantic segmentation. Results show the significant performance gain of our ActiveFT in comparison with baselines.

Our contributions are summarized as follows:
\begin{itemize}
    \item To our best knowledge, we are the first to identify the gap of data selection for annotation and supervised finetuning in the pretraining-finetuning paradigm, which can cause inefficient use of annotation budgets as also verified in our empirical study. Meanwhile, we formulate a new task called \textit{active finetuning} to fill in this gap.
    \item We propose a novel method, ActiveFT, to deal with the active finetuning task through parametric model optimization which theoretically reduces the earth mover's distance (EMD) between the distributions of the selected subset and entire unlabeled pool. To our best knowledge, we are the first to directly optimize samples to be selected in the continuous space for data selection tasks. 
    \item We apply ActiveFT to popular public datasets, achieving leading performance on both classification and segmentation tasks. In particular, our ablation study results justify the design of our method to fill in the data selection gap in the pretraining-finetuning paradigm. The source code will be made public available.
\end{itemize}

\section{Related Work}
\paragraph{Unsupervised Learning} aims to learn the feature representation without the participation of labels. Both contrastive and generative methods achieve great success in this field. Contrastive methods model the similarity and dissimilarity between different input samples. Some early work resorts to a large batch size \cite{chen2020simple} or memory bank \cite{wu2018unsupervised,he2020momentum} to include enough negative samples in each iteration. Challenging the necessity of negative samples, some following study tries to train the network only with positive samples. To this end, they introduce the momentum encoder \cite{grill2020bootstrap}, clustering strategy \cite{caron2020unsupervised}, or stop-gradient operation \cite{chen2021exploring} into contrastive learning frameworks. Based on the success of prior arts, \cite{caron2021emerging,chen2021empirical} succeed in transplanting contrastive learning to vision transformers \cite{dosovitskiy2020image}. Some recent research \cite{he2021masked,bao2021beit,wei2022masked,huang2022contrastive} explores generative methods that predict the missing content inside input samples, also achieving promising performance over vision transformers.

For both kinds of methods, prior research has well investigated their positive roles in downstream supervised finetuning. Of particular interest, they can bring significant performance gain in semi-supervised learning settings, where only a small part (\textit{e.g.} $1\%$) of data samples are annotated.

\paragraph{Active Learning} selects useful samples to fill up the limited annotation budget most beneficial for model training. Despite the existence of \textit{query-synthesizing} \cite{mahapatra2018efficient,mayer2020adversarial,zhu2017generative} and \textit{stream-based} \cite{fang2017learning,narr2016stream} methods, current mainstream approaches are \textit{pool-based}. Given an unlabeled data pool, the pool-based algorithms choose a part of samples for annotation. There are two different selection criteria: \textit{uncertainty} \cite{MIAOD2021,liu2021influence,beluch2018power} and \textit{diversity} \cite{sener2017active,sinha2019variational,agarwal2020contextual}. Based on the uncertainty inside model prediction, the algorithm can select the most difficult data samples. Early work estimates the uncertainty with various heuristics such as posterior probability \cite{lewis1994heterogeneous,wang2016cost}, entropy \cite{joshi2009multi,luo2013latent}, and classification margin \cite{tong2001support}. Some following research directly measures the uncertainty by estimating the training loss \cite{yoo2019learning,huang2021semi} or influence on model performance \cite{liu2021influence,freytag2014selecting} of each sample. Many other algorithms focus on the diversity of selected samples so that the distribution of this selected subset could become close to the original unlabeled pool. To be specific, Sener and Savarese \cite{sener2017active} theoretically formulate the data selection process as a k-Center problem and proposes a CoreSet algorithm. Agarwal et al. \cite{agarwal2020contextual} replace the Euclidean distance with context-aware KL-divergence. Sinha et al. \cite{sinha2019variational} train an adversarial network to discriminate labeled and unlabeled samples. Previous work~\cite{sener2017active,mahmood2021low} also tries to formulate active learning as an optimization problem. They typically pay attention to the discrete space, since it trivially matches the sample distribution inside a dataset. However, discrete optimization problem tends to be much more difficult to solve than continuous problems. Some recent efforts also pay attention to the combination between active learning and unsupervised learning. For example, Yi et al. \cite{yi2022pt4al} guides the data selection with self-supervised learning loss, but their methods only work for some very simple pretext task (\textit{e.g.} colorization, rotation). 

Most above active learning algorithms are designed for from-scratch training. Prior research \cite{bengar2021reducing} reveals their negative effect in finetuning after unsupervised pretraining.

\begin{figure}[t!]
    \centering
    \includegraphics[width=\linewidth]{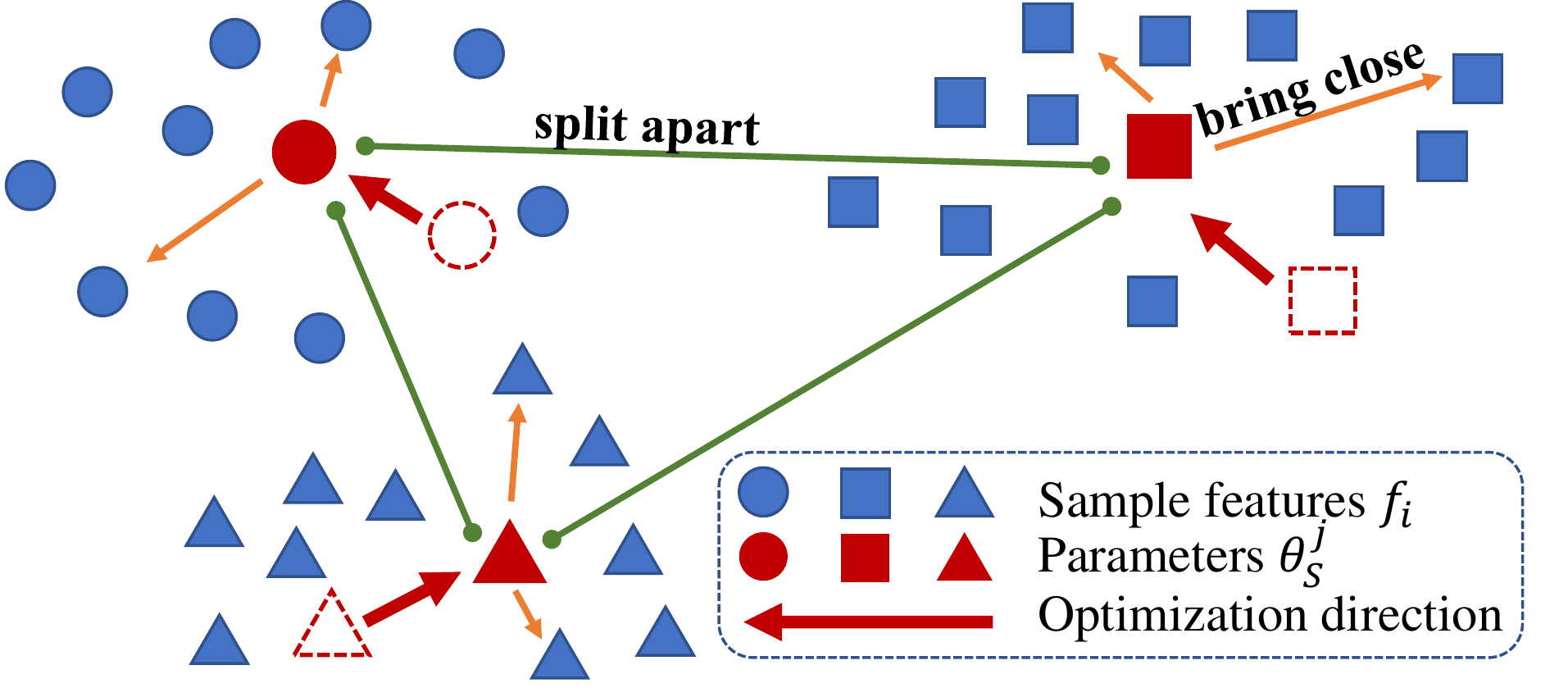}
    \caption{\textbf{Parametric Model Optimization Process:} By optimizing the loss in Eq.~\ref{eq:loss}, each parameter $\theta_{\mathcal{S}}^j$ is appealed by nearby sample features (\textcolor{orange}{orange} in the figure, Eq.~\ref{eq:dterm}) and repelled by other parameters $\theta_{\mathcal{S}}^k,k\neq j$ (\textcolor{colivegreen}{green} in the figure,  Eq.~\ref{eq:rterm}).}
    \label{fig:method}
\end{figure}

\section{Methodology}
We first formulate this new task called \textit{active finetuning} in Sec.~\ref{sec:formulation}. Our novel method, ActiveFT, to solve this problem based on continuous space optimization is proposed in Sec.~\ref{sec:model}. Afterward, we elaborate on how to optimize this model by minimizing the loss function in Sec.~\ref{sec:optimization}. An illustration of our method is shown in Fig.~\ref{fig:method}. We also clarify the correlation between our method and earth mover's distance in Sec.~\ref{sec:math}. Finally, the implementation of this method to deep learning model is explained in Sec.~\ref{sec:implementation}.

\subsection{Formulation of Active Finetuning Task}
\label{sec:formulation}
We formally define the active finetuning task. As is demonstrated in Fig.~\ref{fig:paradigm}, a deep neural network model $f(\cdot;w_0):\mathcal{X}\rightarrow \mathbb{R}^C$ with pretrained weight $w_0$ is given, where $\mathcal{X}$ is the data space and $\mathbb{R}^C$ is the normalized high-dimensional feature space. We also have access to a large unlabeled data pool $\mathcal{P}^u=\{\mathbf{x}_i\}_{i\in [N]}\sim p_u$ inside data space $\mathcal{X}$ with distribution $p_u$, where $[N]=\{1,2,\dots,N\}$. The subset $\mathcal{P}^u_{\mathcal{S}}$ for supervised finetuning is selected from $\mathcal{P}^u$. It is worth noting that $f(\cdot;w_0)$ can be pretrained either on $\mathcal{P}^u$ or other data sources, \textit{e.g.} pretrained on ImageNet-1k \cite{russakovsky2015imagenet} and finetuned on a subset of CIFAR-10 \cite{krizhevsky2009learning}.

In the active finetuning task, we should design a sampling strategy $\mathcal{S}=\{s_j\in[N]\}_{j\in[B]}$ to select a subset $\mathcal{P}_{\mathcal{S}}^u=\{\mathbf{x}_{s_j}\}_{j\in[B]}\subset \mathcal{P}^u$ from $\mathcal{P}^u$, where $B$ is the annotation budget size for supervised finetuning. The model would have access to the labels $\{\mathbf{y}_{s_j}\}_{j\in[B]}\subset\mathcal{Y}$ of this subset through the oracle, obtaining a labeled data pool $\mathcal{P}^l_\mathcal{S}=\{\mathbf{x}_{s_j},\mathbf{y}_{s_j}\}_{j\in[B]}$, where $\mathcal{Y}$ is the label space. Afterward, the model $f$ is finetuned on $\mathcal{P}^l_\mathcal{S}$ supervisedly and the model parameter is updated to $w_{\mathcal{S}}$ after the finetuning. The goal of active finetuning is to find the sampling strategy $\mathcal{S}_{opt}$ minimizing the expected model error $error(f(\mathbf{x};w_{\mathcal{S}}),\mathbf{y})$.
\begin{equation}
    \mathcal{S}_{opt}=\arg\min_{\mathcal{S}}\underset{\mathbf{x},\mathbf{y}\in \mathcal{X}\times \mathcal{Y}}{E}\left[error(f(\mathbf{x};w_{\mathcal{S}}),\mathbf{y})\right]
\end{equation}

Our active finetuning is different from traditional active learning in: 1) We have access to the pretrained model $f(\cdot;w_0)$, which will be finetuned, before data selection. 2) The selected samples are applied to the finetuning of the pretrained model $f(\cdot;w_0)$ instead of from-scratch training. 3) The sampled subset size $|\mathcal{P}_{\mathcal{S}}^l|$ is relatively small, less than $10\%$ in most cases. 4) We have no access to any labels such as a random initial labeled set before data selection.

\subsection{Data Selection with Parametric Model}
\label{sec:model}
We select samples under the guidance two basic intuitions: \textit{1) bringing close the distributions between the selected subset $\mathcal{P}_{\mathcal{S}}^u$ and the original pool $\mathcal{P}^u\sim p_u$. 2) maintaining the diversity of $\mathcal{P}_{\mathcal{S}}^u$}. The former ensures the model finetuned on the subset performs similarly with that trained on the full set, while the latter allows the subset to cover corner cases in the full set. In comparison to distribution $p_u(\mathbf{x})$ in the data space, it is more feasible to work on its corresponding distribution $p_{f_u}(\mathbf{f})$ in the feature space. Through the agency of pretrained model $f(\cdot;w_0)$, we map each data sample $\mathbf{x}_i$ to the high dimensional feature space as $\mathbf{f}_i=f(\mathbf{x}_i;w_0)$, where $\mathbf{f}_i$ is the \textit{normalized} feature of $\mathbf{x}_i$. As a result, we can derive the pool $\mathcal{F}^u=\{\mathbf{f}_i\}_{i\in[N]}$ from $\mathcal{P}^u$ and corresponding distribution $p_{f_u}$ of $\mathcal{F}^u$. 

Similarly, the feature pool $\mathcal{F}_{\mathcal{S}}^u$ is also associated with the selected data subset $\mathcal{P}_{\mathcal{S}}^u$. We define the corresponding distribution over $\mathcal{F}_{\mathcal{S}}^u$ in the feature space as $p_{f_{\mathcal{S}}}$. Our goal is to find the optimal selection strategy $\mathcal{S}$ as follows.
\begin{equation}
    \mathcal{S}_{opt}=\arg\min_{\mathcal{S}}D(p_{f_u},p_{f_{\mathcal{S}}})-\lambda R(\mathcal{F}_{\mathcal{S}}^u)
    \label{eq:goal}
\end{equation}
where $D(\cdot,\cdot)$ is some distance metrics between distributions, $R(\cdot)$ is to measure the diversity of a set, and $\lambda$ is a scale to balance these two terms. The first term aims to bring close these two distributions $p_{f_u},p_{f_{\mathcal{S}}}$ while the second term is to ensure the diversity of subset.

Unfortunately, it is difficult to directly optimize the \textit{discrete} selection strategy $\mathcal{S}$, so we alternatively model $p_{f_{\mathcal{S}}}$ with $p_{\theta_{\mathcal{S}}}$, where $\theta_{\mathcal{S}}=\{\theta_\mathcal{S}^j\}_{j\in[B]}$ are the \textit{continuous} parameters and $B$ is the annotation budget size. Each $\theta_{\mathcal{S}}^j$  after optimization corresponds to the feature of a selected sample $\mathbf{f}_{s_j}$. We would find $\mathbf{f}_{s_j}$ closest to each $\theta_{\mathcal{S}}^j$ after optimization to determine the selection strategy $\mathcal{S}$. Therefore, our goal in Eq.~\ref{eq:goal} is written as follows.
\begin{equation}
    \theta_{\mathcal{S},opt}=\arg\min_{\theta_{\mathcal{S}}}D(p_{f_u},p_{\theta_{\mathcal{S}}})-\lambda R(\theta_{\mathcal{S}})\ s.t.\ ||\theta_{\mathcal{S}}^j||_2=1
    \label{eq:distrib_dist}
\end{equation}
The difference between extracted sample features $\mathcal{F}_{\mathcal{S}}^u=\{\mathbf{f}_{s_i}\}$ and our define parameters $\theta_{\mathcal{S}}=\{\theta_\mathcal{S}^j\}$ is that $\mathbf{f}_{s_i}$ is a discrete feature corresponding to a sample in the dataset while $\theta_\mathcal{S}^j$ is continuous in the feature space.

\subsection{Parametric Model Optimization}
\label{sec:optimization}
In the parametric model $p_{\theta_{\mathcal{S}}}$, the distribution is represented by $B$ parameters $\{\theta_{\mathcal{S}}^j\}_{j\in[B]}$. We consider it as a mixture model with $B$ components in Eq.~\ref{eq:mixture}.
\begin{equation}
    p_{\theta_{\mathcal{S}}}(\mathbf{f})=\sum_{j=1}^B\phi_{j}p(\mathbf{f}|\theta_{\mathcal{S}}^j)
    \label{eq:mixture}
\end{equation}
where $\phi_{j}$ is the mixture weight or prior probability $p(\theta_{\mathcal{S}}^j)$ of the $j$-th component, satisfying $\sum_{j=1}^B\phi_j=1$. Since $\mathbf{f}$ and $\theta_{\mathcal{S}}^j$ both lie in the feature space, we formulate the distribution of each component based on their similarity as Eq.~\ref{eq:component}.
\begin{equation}
    p(\mathbf{f}|\theta_{\mathcal{S}}^j)=\frac{\exp(sim(\mathbf{f},\theta_{\mathcal{S}}^j)/\tau)}{Z_j}
    \label{eq:component}
\end{equation}
where $Z_j$ is a normalizing constant, $sim(\cdot,\cdot)$ is a similarity metric, and $\tau$ is the temperature scale. We follow the protocol in  \cite{wu2018unsupervised,caron2021emerging} to apply the cosine similarity between normalized features as the metric $sim(\mathbf{f}_1,\mathbf{f}_2)=\mathbf{f}_1^\top\mathbf{f}_2,||\mathbf{f}_1||_2=||\mathbf{f}_2||_2=1$ and set the temperature $\tau=0.07$ \cite{wu2018unsupervised,caron2021emerging} throughout the paper. For each $\mathbf{f}_i\in \mathcal{F}^u$, there exists a $\theta_{\mathcal{S}}^{c_i}$ most similar (and closest) to $\mathbf{f}_i$, \textit{i.e.}
\begin{equation}
    c_i=\arg\max_{j\in[B]}sim(\mathbf{f}_i,\theta_{\mathcal{S}}^j)
    \label{eq:closest}
\end{equation}
where we keep updating $c_i$ in the optimization process.

\begin{figure}[t!]
    \centering
    \includegraphics[width=0.45\linewidth]{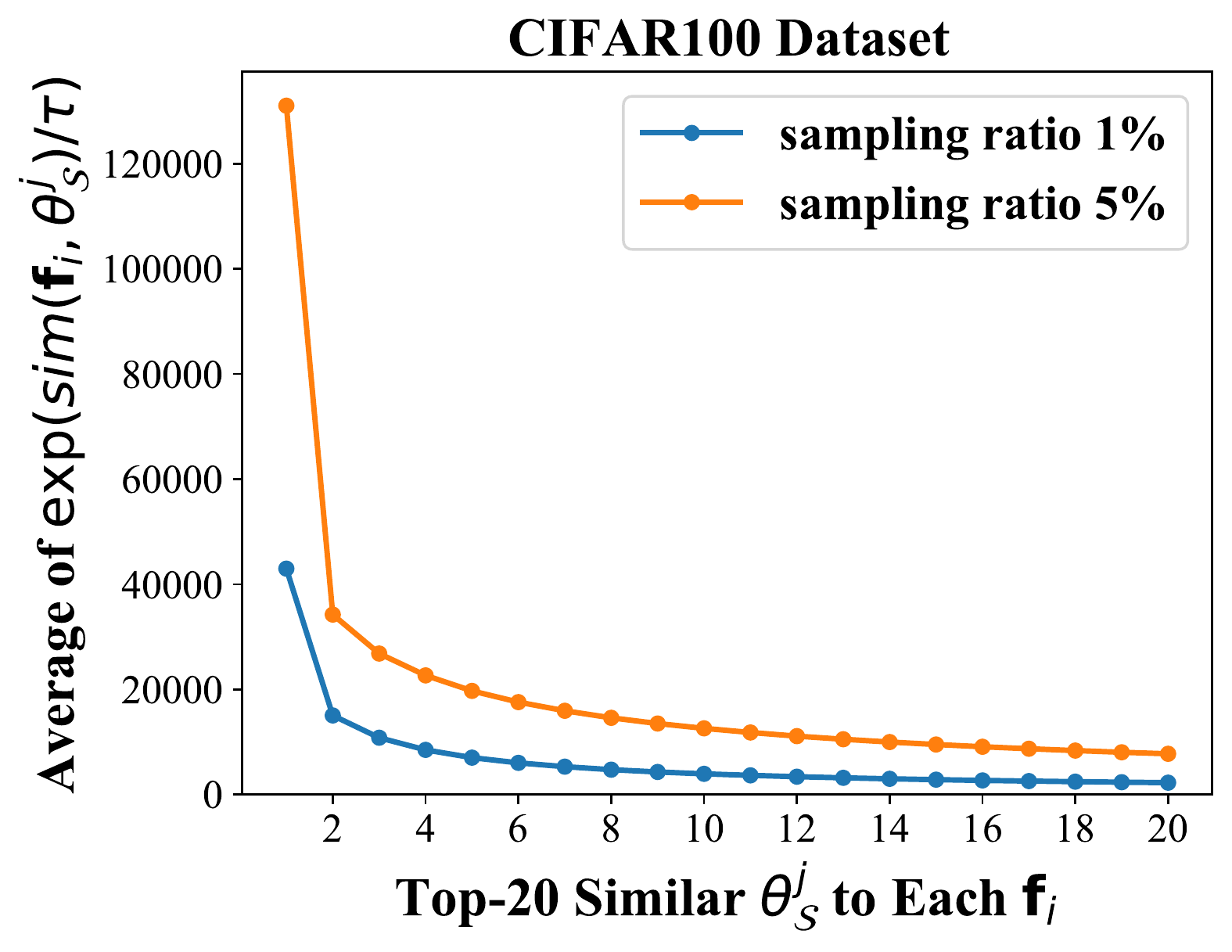}
    \includegraphics[width=0.45\linewidth]{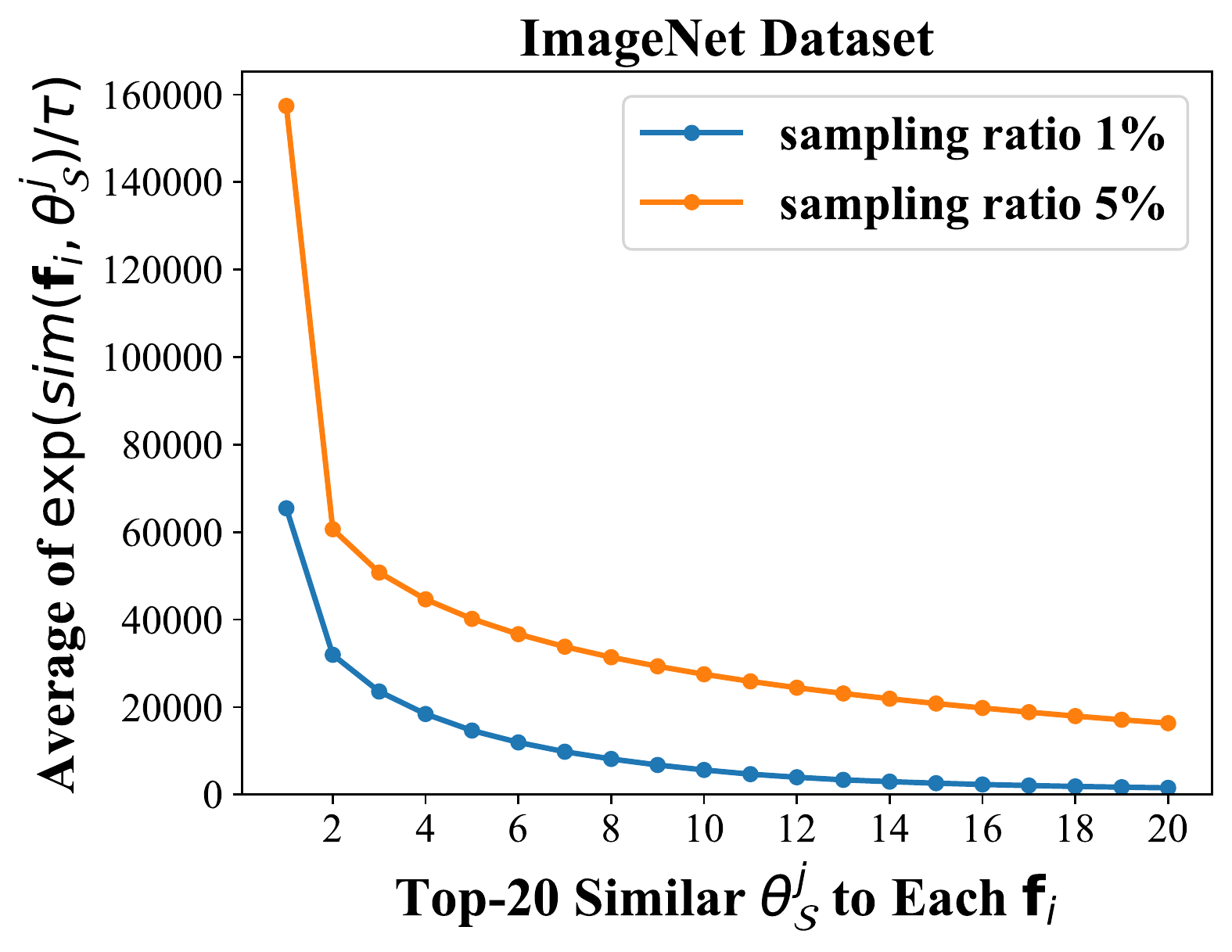}
    \caption{\textbf{Similarity between Features and Parameters:} On CIFAR100 and ImageNet, we find the Top-20 most similar parameters $\theta_{\mathcal{S}}^j$ with each sample feature $\mathbf{f}_i$, and calculate the average exponential similarity $E_{i\in[N]}[\exp(sim(\mathbf{f}_i,\theta_{\mathcal{S}}^j)/\tau]$. Here $\theta_{\mathcal{S}}=\{\theta_{\mathcal{S}}^j\}_{j\in[B]}$ is randomly sampled following the distribution $p_{f_u}$. The model $f(\cdot;w_0)$ is DeiT-Small \cite{touvron2021training} pretrained on ImageNet \cite{russakovsky2015imagenet} with DINO framework \cite{caron2021emerging}. The results verify Assumption~\ref{eq:large} that the Top-1 similarity is significantly larger than others.}
    \label{fig:large}
\end{figure}

Since there is a very low temperature ($\tau=0.07$), the gap between the exponential similarity $\exp(sim(\mathbf{f}_i,\theta_{\mathbf{S}}^j)/\tau)$ with different $\theta_{\mathbf{S}}^j$ is significant. Therefore, it is safe to make the following \textit{assumption}.
\begin{assumption}
    $\forall i\in[N],j\in[B]$, if $\tau$ is small, the following far-more-than relationship holds that
    $$\exp(sim(\mathbf{f}_i,\theta_{\mathcal{S}}^{c_i})/\tau)\gg  \exp(sim(\mathbf{f}_i,\theta_{\mathcal{S}}^j)/\tau),j\neq c_i$$
    \label{eq:large}
    \vspace{-20pt}
\end{assumption}
Although it is hard to mathematically prove, this assumption is empirically verified by the results in Fig.~\ref{fig:large}. In another word, $p(\mathbf{f}_i|\theta_{\mathcal{S}}^{c_i})\gg p(\mathbf{f}_i|\theta_{\mathcal{S}}^{j}),j\neq c_i,j\in[B]$. We can approximate the parametric model for $\mathbf{f}_i\in \mathcal{F}^u$ in Eq.~\ref{eq:mixture}.
\begin{equation}
\begin{aligned}
     p_{\theta_{\mathcal{S}}}(\mathbf{f}_i) &\approx 
    \phi_{c_i}p(\mathbf{f}_i|\theta^{c_i}_{\mathcal{S}})\\ &=\frac{\exp(sim(\mathbf{f}_i,\theta_{\mathcal{S}}^{c_i})/\tau)}{Z_{c_i}/\phi_{c_i}}\\ &=\frac{\exp(sim(\mathbf{f}_i,\theta_{\mathcal{S}}^{c_i})/\tau)}{\tilde{Z}_{c_i}}\\
\end{aligned}
\label{eq:approx}
\end{equation}
where $\tilde{Z}_{c_i}=Z_{c_i}/\phi_{c_i}$ is a new normalizing constant. We can derive $p_{\theta_{\mathcal{S}}}(\mathbf{f}_i)\propto \exp(sim(\mathbf{f}_i,\theta_{\mathcal{S}}^{c_i})/\tau)$ from Eq.~\ref{eq:approx}.

The two distributions $p_{f_u},p_{\theta_{\mathcal{S}}}$ can be brought close by minimizing the KL-divergence.
\begin{equation}
\begin{aligned}
    KL(p_{f_u}|p_{\theta_{\mathcal{S}}})&=\sum_{\mathbf{f}_i\in \mathcal{F}^u}p_{f_u}(\mathbf{f}_i)\log\frac{p_{f_u}(\mathbf{f}_i)}{p_{\theta_{\mathcal{S}}}(\mathbf{f}_i)}\\
    &=\underset{\mathbf{f}_i\in \mathcal{F}^u}{E}\left[\log p_{f_u}(\mathbf{f}_i)\right]-\underset{\mathbf{f}_i\in \mathcal{F}^u}{E}\left[\log p_{\theta_{\mathcal{S}}}(\mathbf{f}_i)\right]
\end{aligned}
\label{eq:kl}
\end{equation}
where the first term $\underset{\mathbf{f}_i\in \mathcal{F}^u}{E}\left[\log p_{f_u}(\mathbf{f}_i)\right]$ is a constant without the parameter $\theta_{\mathcal{S}}$. Then, minimizing the KL-divergence $KL(p_{f_u}|p_{\theta_{\mathcal{S}}})$ equals to maximizing the second term $\underset{\mathbf{f}_i\in \mathcal{F}^u}{E}\left[\log p_{\theta_{\mathcal{S}}}(\mathbf{f}_i)\right]$, and according to Eq.~\ref{eq:approx}, it is also equivalent with maximizing $\underset{\mathbf{f}_i\in \mathcal{F}^u}{E}\left[\log \exp(sim(\mathbf{f}_i,\theta_{\mathcal{S}}^{c_i})/\tau)\right]=\underset{\mathbf{f}_i\in \mathcal{F}^u}{E}\left[sim(\mathbf{f}_i,\theta_{\mathcal{S}}^{c_i})/\tau\right]$. Therefore, we derive the first term in Eq.~\ref{eq:distrib_dist} as follows.
\begin{equation}
    D(p_{f_u},p_{\theta_{\mathcal{S}}})=-\underset{\mathbf{f}_i\in \mathcal{F}^u}{E}\left[sim(\mathbf{f}_i,\theta_{\mathcal{S}}^{c_i})/\tau\right]
    \label{eq:dterm}
\end{equation}


However, directly carrying out this optimization leads to a severe \textit{collapse} problem, \textit{i.e.} most $\theta_{\mathcal{S}}^j,j\in[B]$ converge to the same position with the highest density of $\mathbf{f}_i,i\in[N]$, losing the diversity inside the selected data. To this end, as shown in Eq.~\ref{eq:goal}, we introduce an extra regularization term to ensure the diversity of selected subset. Without bells and whistles, this regularization is implemented by minimizing the similarity between selected samples. We also add an exponential operation to make the optimization process more stable, otherwise some $\theta_{\mathcal{S}}^j$ may become outliers.
\begin{equation}
    R(\theta_{\mathcal{S}})=-\underset{j\in[B]}{E}\left[\log\sum_{k\neq j,k\in[B]}\exp\left(sim(\theta_{\mathcal{S}}^j,\theta_{\mathcal{S}}^k)/\tau\right)\right]
    \label{eq:rterm}
\end{equation}

At this point, we are able to solve Eq.~\ref{eq:distrib_dist} by optimizing the following loss function continuously.
\begin{equation}
\label{eq:loss}
        \resizebox{\linewidth}{!}{
        \begin{math}
    \begin{aligned} 
        L&=D(p_{f_u},p_{\theta_{\mathcal{S}}})-\lambda\cdot  R(\theta_{\mathcal{S}})\\
        &=-\underset{\mathbf{f}_i\in \mathcal{F}^u}{E}\left[sim(\mathbf{f}_i,\theta_{\mathcal{S}}^{c_i})/\tau\right]+\underset{j\in[B]}{E}\left[\log\sum_{k\neq j,k\in[B]}\exp\left(sim(\theta_{\mathcal{S}}^j,\theta_{\mathcal{S}}^k)/\tau\right)\right]
    \end{aligned}
    \end{math}
    }
\end{equation}
where the balance weight $\lambda$ is empirically set as $1$.

Finally, we directly optimize the loss function in Eq.~\ref{eq:loss} by gradient descent. When the optimization is finished, we find feature $\{\mathbf{f}_{s_j}\}_{j\in[B]}$ with the highest similarity to $\theta_{\mathcal{S}}^j$.
\begin{equation}
    \mathbf{f}_{s_j}=\arg\max_{\mathbf{f}_k\in\mathcal{F}^u}sim(\mathbf{f}_k,\theta_{\mathcal{S}}^j)
    \label{eq:sim}
\end{equation}
The corresponding data samples $\{\mathbf{x}_{s_j}\}_{j\in[B]}$ are selected as the subset $\mathcal{P}_{\mathcal{S}}^u$ with selection strategy $\mathcal{S}=\{s_j\}_{j\in[B]}$.

\subsection{Relation to Earth Mover's Distance}
\label{sec:math}
In this part, we show that optimizing the loss in Eq.~\ref{eq:loss} is actually minimizing the earth mover's distance between the distributions of selected subset and full set. This justifies that our optimization in the continuous space is equivalent with bringing close the distribution gap in the discrete data sample space.

After the optimization, we get the features $\mathbf{f}_{s_j}$ of selected samples. We deliberately assign the discrete probability distribution $p_{f_S}$ as Eq.~\ref{eq:dist_fu}.
\begin{equation}
    p_{f_S}(\mathbf{f}_{s_j})=\frac{|C_j|}{N},C_j=\{\mathbf{f}_i|c_i=j\},\mathbf{f}_{s_j}\in\mathcal{F}^u_{\mathcal{S}}
    \label{eq:dist_fu}
\end{equation}
where $C_j$ is the set of features closest to $f_{s_j}$ with $c_i$ defined in Eq.~\ref{eq:closest}. The distribution $p_{f_u}$ is modeled as a uniform distribution over $\mathcal{F}^u$, \textit{i.e.} $p_{f_u}(\mathbf{f}_i)=\frac{1}{N},\mathbf{f}_i\in\mathcal{F}^u$.

The earth mover's distance (EMD) between $p_{f_u},p_{f_S}$ is written as~\cite{levina2001earth}:
\begin{equation}
    EMD(p_{f_u},p_{f_S})=\underset{\gamma\in\Pi(p_{f_u},p_{f_S})}{inf}\underset{(\mathbf{f}_i,\mathbf{f}_{s_j})\sim \gamma}{E}\left[||\mathbf{f}_i-\mathbf{f}_{s_j}||_2\right]
    \label{eq:wdist}
\end{equation}
where $\Pi(p_{f_u},p_{f_S})$ is the set of all possible joint distributions whose marginals are $p_{f_u}$ and $p_{f_S}$. It is intuitive to come up with the infimum, \textit{i.e.} each $\mathbf{f}_i\sim p_{f_u}$ transports to their closest $\mathbf{f}_{s_j}\sim p_{f_{\mathcal{S}}}$. The detailed derivation is in the appendix.
\begin{equation}
    \gamma_{f_u,f_S}(\mathbf{f}_i,\mathbf{f}_{s_j})=\begin{cases}
    \frac{1}{N} & \mathbf{f}_i\in \mathcal{F}^u,\mathbf{f}_{s_j}\in\mathcal{F}^u_{\mathcal{S}},c_i=j\\
    0 & otherwise\\
    \end{cases}
    \label{eq:transport}
\end{equation}
 In this case, the distance in Eq.~\ref{eq:wdist} becomes 
\begin{equation}
\begin{aligned}
    EMD(p_{f_u},p_{f_S})&=\underset{(\mathbf{f}_i,\mathbf{f}_{s_{c_i}})\sim \gamma}{E}\left[||\mathbf{f}_i-\mathbf{f}_{s_{c_i}}||_2\right]\\
    &=\frac{1}{N}\sum_{i=1}^N\left[\sqrt{2-2sim(\mathbf{f}_i,\mathbf{f}_{s_{c_i}})}\right]
\end{aligned}
\label{eq:wdistres}
\end{equation}

In Eq.~\ref{eq:loss}, we minimize $-sim(\mathbf{f}_i,\theta_{\mathcal{S}}^{c_i})$, and $\mathbf{f}_{s_{c_i}}$ is set as the closest $\mathbf{f}_k\in\mathcal{F}^u$ to $\theta_{\mathcal{S}}^{c_i}$ in Eq.~\ref{eq:sim} after optimization. Then, the distance in Eq.~\ref{eq:wdistres} is actually also minimized. Therefore, our optimization method in Sec.~\ref{sec:optimization} is equivalent with reducing the earth mover's distance between the distributions of the original unlabeled pool and selected subset.

\begin{algorithm}[ht!]
    \caption{\textbf{Pseudo-code for ActiveFT}}
    \label{alg:optimization}
    \KwInput{Unlabeled data pool $\{\mathbf{x}_i\}_{i\in[N]}$, pretrained model $f(\cdot;w_0)$, annotation budget $B$, iteration number $T$ for optimization}
    \KwOutput{Optimal selection strategy $\mathcal{S}=\{s_j\in[N]\}_{j\in[B]}$}
    
    \For{$i\in[N]$}{
        $\mathbf{f}_i=f(\mathbf{x}_i;w_0)$\\
    }
    \tcc{Construct $\mathcal{F}^u=\{\mathbf{f}_i\}_{i\in[N]}$ based on $\mathcal{P}^u$, normalized to $||\mathbf{f}_i||_2=1$}
    
    Uniformly random sample $\{s_j^0\in[N]\}_{j\in [B]}$, and initialize $\theta_{\mathcal{S}}^j=\mathbf{f}_{s_j^0}$\\
    \tcc{Initialize the parameter $\theta_{\mathcal{S}}=\{\theta_{\mathcal{S}}^j\}_{j\in[B]}$ based on $\mathcal{F}^u$}
    
    \For{$iter\in[T]$}{
        Calculate the similarity between $\{\mathbf{f}_i\}_{i\in[N]}$ and $\{\theta_{\mathcal{S}}^j\}_{j\in[B]}$: $Sim_{i,j}=\mathbf{f}_{i}^\top\theta_{\mathcal{S}}^j/\tau$\\
        $MaxSim_i=\max_{j\in[B]}Sim_{i,j}=Sim_{i,c_i}$\\
        \tcc{The Top-1 similarity between $\mathbf{f}_{i}$ and $\theta_{\mathcal{S}}^j,j\in[B]$}
        Calculate the similarity between $\theta_{\mathcal{S}}^j$ and $\theta_{\mathcal{S}}^k,k\neq j$ for regularization: $RegSim_{j,k}=\exp({\theta_{\mathcal{S}}^j}^\top\theta_{\mathcal{S}}^k/\tau),k\neq j$\\
        $Loss=-\frac{1}{N}\sum_{i\in[N]}MaxSim_i+\frac{1}{B}\sum_{j\in[B]}\log\left(\sum_{k\neq j}RegSim_{j,k}\right)$\\
        \tcc{Calculate the loss function in Eq.~\ref{eq:loss}}
        $\theta_{\mathcal{S}}=\theta_{\mathcal{S}}-lr\cdot \bigtriangledown_{\theta_{\mathcal{S}}}Loss$\\
        \tcc{Optimize the parameter through gradient descent}
        $\theta_{\mathcal{S}}^j=\theta_{\mathcal{S}}^j/||\theta_{\mathcal{S}}^j||_2,j\in[B]$\\
        \tcc{Normalize the parameters to ensure $||\theta_{\mathcal{S}}^j||_2=1$}
    }
    Find $\mathbf{f}_{s_j}$ closest to $\theta_{\mathcal{S}}^j$: $s_j=\arg\max_{k\in[N]} \mathbf{f}_k^\top\theta_{\mathcal{S}}^j$ for each $j\in[B]$\\
    Return the selection strategy $\mathcal{S}=\{s_j\}_{j\in[B]}$\\
\end{algorithm}

\subsection{Implementation as a Learning Model}
\label{sec:implementation}
Alg.~\ref{alg:optimization} shows how to implement this method to deep learning models. Given a pretrained model, for each image sample $\mathbf{x}_i\in\mathcal{P}^u$, we extract the last layer [CLS] token feature in the transformer model or global pooling feature in the convolutional model, which is normalized as the high-dimensional feature $\mathbf{f}_i=f(\mathbf{x}_i;w_0)$. Before the optimization process, the parameter $\theta_{\mathcal{S}}$ is initialized by uniformly sampling $\theta_{\mathcal{S}}^j,j\in[B]$ at random from the feature pool $\mathcal{F}^u=\{\mathbf{f}_i\}_{i\in[N]}$. If $|\mathcal{F}^u|$ is extremely large, we would randomly select $M$ elements from $\mathcal{F}^u$ (\textit{e.g.} $M$=100,000 for ImageNet dataset) for the each training iteration of our parametric model. In each iteration, we calculate the similarity between sample features and parameters, then update $c_i$ in Eq.~\ref{eq:closest} for each $\mathbf{f}_i$ and positive feature set $\{\mathbf{f}_i|c_i=j\}$ for each $\theta_{\mathcal{S}}^j$. Afterwards, we can compute the loss function in Eq.~\ref{eq:loss} and update the parameters $\theta_{\mathcal{S}}$ by gradient descent. When the optimization process is finished, we find the sample feature $\mathbf{f}_{s_j}$ most similar to each parameter $\theta_{\mathcal{S}}^j$ (Eq.~\ref{eq:sim}). Those corresponding samples $\{\mathbf{x}_{s_j}\}_{j\in[B]}$ are selected for annotation for the following supervised finetuning.

\section{Experiments}
Our method is evaluated on image classification (Sec.~\ref{sec:classification}) and semantic segmentation (Sec.~\ref{sec:segmentation}) tasks. The performance is compared with some baselines and traditional active learning algorithms. We make some qualitative and quantitative analyses of our method in Sec.~\ref{sec:analysis}. Finally, the roles of different modules inside our method are examined in Sec.~\ref{sec:ablation}. Experiments are run on GeForce RTX 3090 (24GB) and AMD Ryzen Threadripper 3970X 32-Core Processor.

\subsection{Image Classification Results}
\label{sec:classification}
\textbf{Datasets and Benchmarks} For classification task, we choose three classical datasets CIFAR10, CIFAR100 \cite{krizhevsky2009learning}, and ImageNet-1k \cite{russakovsky2015imagenet} for experiments. CIFAR10 and CIFAR100 contain 60,000 images of 32x32 scale with 10 and 100 categories separately, among which 50,000 images are in the training set and 10,000 images are for test. ImageNet-1k is a large-scale dataset spans 1000 classes, containing 1,281,167 training images and 50,000 validation images. Their training sets are considered as the unlabeled pool $\mathcal{P}^u$ for selection.  We evaluate the performance with the \textit{Top-1 Accuracy} metric. 

\textbf{Implementation Details}
Our method is agnostic with pretraining frameworks and networks. We apply DeiT-Small \cite{touvron2021training} pretrained with DINO \cite{caron2021emerging} framework on ImageNet-1k \cite{russakovsky2015imagenet} in the experiments for its verified popularity and effectiveness. We also attempt other architectures in Sec.~\ref{sec:analysis}.  For all three datasets, we resize images to 224x224 consistent with the pretraining for both data selection and supervised finetuning. In the data selection process, the parameters $\theta_{\mathcal{S}}$ are optimized with Adam \cite{kingma2014adam} optimizer (learning rate 1e-3) until convergence. The experiment details of supervised finetuning are available in the \textit{supplementary materials}.

\textbf{Baselines} We compare our method with eight counterparts including the following three baselines and five traditional active learning methods.
\begin{enumerate}
    \item \textbf{Random:} The samples for annotation are selected uniformly at random. 
    \item \textbf{FDS:} \textit{a.k.a} K-Center-Greedy algorithm. It selects the next sample feature farthest from current selections. Proved in \cite{sener2017active}, it minimizes the gap between the expected loss over the entire pool and the selected subset. In accordance with the pretraining process \cite{caron2021emerging}, we apply cosine distance as the distance metric. 
    \item \textbf{K-Means:} We conduct K-Means over the feature pool $\mathcal{F}^u$ and choose samples closest to the centers. $K$ equals to the budget size $B$.
\end{enumerate}
We transplant five active learning algorithms CoreSet \cite{sener2017active}, VAAL \cite{sinha2019variational}, LearnLoss \cite{yoo2019learning}, TA-VAAL\cite{kim2021task} and ALFA-Mix\cite{parvaneh2022active} to our active finetuning task. The former three are classical and the latter two are newer, all  equipped with open-source codes. We refer readers to the appendix for transplantation details.

\textbf{Results and Comparison}
We average our results over three independent runs. The results are shown in Tab.~\ref{tab:classification}. Traditional active learning methods typically fail in the pretraining-finetuning paradigm, which is consistent with the results in \cite{bengar2021reducing}. In contrast, our ActiveFT outperforms counterparts on all three datasets with different sampling ratios. On each dataset, the performance gain is especially significant when the sampling ratio is low, since our method can select the most representative samples. This phenomenon is of great practical use because the sampling number for supervised finetuning is usually much smaller than the pool size to save the annotation cost.

\begin{table*}[t!]
\caption{\textbf{Image Classification Results:} Experiments are conducted on natural images with different sampling ratios. We report the mean and std over three trials. Explanation of N/A results (``-") is in our appendix.} 
\label{tab:classification}
\centering
\resizebox{\linewidth}{!}{
\begin{tabular}{l|ccc|cccc|ccc}
     \toprule
     \multirow{2}{*}{\textbf{Methods}} & \multicolumn{3}{c}{\textbf{CIFAR10}} & \multicolumn{4}{c}{\textbf{CIFAR100}} & \multicolumn{2}{c}{\textbf{ImageNet}}\\
     & $0.5\%$ & $1\%$ & $2\%$ & $1\%$ & $2\%$ & $5\%$ & $10\%$ & $1\%$ & $5\%$\\
     \midrule
     \textbf{Random} & 77.3$\pm$2.6 &82.2$\pm$1.9 &88.9$\pm$0.4 &14.9$\pm$1.9 &24.3$\pm$2.0 &50.8$\pm$3.4 &69.3$\pm$0.7 & 45.1$\pm$0.8 &64.3$\pm$0.3\\
     \textbf{FDS} & 64.5$\pm$1.5 & 73.2$\pm$1.2 & 81.4$\pm$0.7 & 8.1$\pm$0.6 & 12.8$\pm$0.3 & 16.9$\pm$1.4 & 52.3$\pm$1.9 & 26.7$\pm$0.6 & 55.5$\pm$0.1\\
     \textbf{K-Means} & 83.0$\pm$3.5 & 85.9$\pm$0.8 & 89.6$\pm$0.6& 17.6$\pm$1.1 & 31.9$\pm$0.1 & 42.4$\pm$1.0 & 70.7$\pm$0.3 & - & -\\
     \midrule
     \textbf{CoreSet} \cite{sener2017active} & - & 81.6$\pm$0.3 & 88.4$\pm$0.2 & - & 30.6$\pm$0.4 & 48.3$\pm$0.5 & 62.9$\pm$0.6 & - & 61.7$\pm$0.2\\
     \textbf{VAAL} \cite{sinha2019variational} & -& 80.9$\pm$0.5 & 88.8$\pm$0.3 & - & 24.6$\pm$1.1 & 46.4$\pm$0.8 & 70.1$\pm$0.4&-& 64.0$\pm$0.3\\
     \textbf{LearnLoss} \cite{yoo2019learning} & -& 81.6$\pm$0.6 & 86.7$\pm$0.4 & - & 19.2$\pm$2.2 & 38.2$\pm$2.8 & 65.7$\pm$1.1&- & 63.2$\pm$0.4\\
     \textbf{TA-VAAL}\cite{kim2021task}  & - & 82.6$\pm$0.4 & 88.7$\pm$0.2 & - & 34.7$\pm$0.7 & 46.4$\pm$1.1 & 66.8$\pm$0.5 & - & 64.3$\pm$0.2 \\
     \textbf{ALFA-Mix}\cite{parvaneh2022active} & - & 83.4$\pm$0.3 & 89.6$\pm$0.2 & - & 35.3$\pm$0.8 & 50.4$\pm$0.9 & 69.9$\pm$0.6 & - & 64.5$\pm$0.2 \\
     \midrule
    \textbf{ActiveFT (ours)} & \textbf{85.0}$\pm$0.4 & \textbf{88.2}$\pm$0.4 & \textbf{90.1}$\pm$0.2 & \textbf{26.1}$\pm$2.6&\textbf{40.7}$\pm$0.9&\textbf{54.6}$\pm$2.3&\textbf{71.0}$\pm$0.5 & \textbf{50.1}$\pm$0.3 & \textbf{65.3}$\pm$0.1\\
     \bottomrule
\end{tabular}
}
\end{table*}

\subsection{Semantic Segmentation Results}
\label{sec:segmentation}
\textbf{Datasets and Benchmarks} 
For segmentation task, we apply ADE20k dataset \cite{zhou2017scene}. It contains 20,210 images for training, 2,000 images for validation, and 3,352 images for test. All images have fine-grained labels with 150 semantic classes. The training set is considered as the unlabeled pool $\mathcal{P}^u$ for selection. We evaluate the performance with the \textit{mIoU} metric. 

\textbf{Implementation Details} Same with image classification task, we apply DeiT-Small \cite{touvron2021training} model pretrained with DINO framework \cite{caron2021emerging} for data selection. The images are resized to 224x224 as well. Since the semantic segmentation task relies more on the local information inside images, we concatenate the [CLS] token features with the average features of other tokens as $\mathbf{f}_i$ for data selection. For the segmentation task, Segmenter \cite{strudel2021segmenter} is adopted for finetuning, which is a pure transformer model. We use the same pretrained DeiT-Small \cite{caron2021emerging} as its backbone. The finetuning details are also available in appendix. 

\begin{table}[t]
\caption{\textbf{Semantic Segmentation Results:} experiments are conducted on ADE20k with sampling ratios $5\%,10\%$. Results are averaged over three trials.}
\label{tab:segmentation}
\centering
\resizebox{\linewidth}{!}{
\begin{tabular}{c|ccccc}
     \toprule
     \textbf{Sel. Ratio} & \textbf{Random} & \textbf{FDS}& \textbf{K-Means} & \textbf{ActiveFT (ours)}\\
     \midrule
    $5\%$ & 14.54 & 6.74 & 13.62 &  \textbf{15.37}$\pm$0.11\\
    $10\%$ & 20.27 & 12.65 & 19.12& \textbf{21.60}$\pm$0.40\\
     \bottomrule
\end{tabular}
}
\end{table}

\textbf{Results and Comparison} In Tab.~\ref{tab:segmentation}, we report the performance of our method when choosing $5\%,10\%$ of training samples for finetuning. Our results are averaged over three independent trials. We compare to three baselines same with image classification. The traditional active learning methods are not included due to their failure on image classification. The performance gain of our data selection method is not as significant as the image classification task. This is understandable because semantic segmentation is a fine-grained vision task, focusing on subtle local visual pattern. In this case, it is hard for a global feature to represent all the details in a scene. However, despite the difficulty, our method still shows notable superiority in comparison with other baselines, reflecting the generality of our method to different tasks. 

\begin{table}[t]
\caption{\textbf{Data Selection Efficiency:} We compare the time cost to select different percentages of samples from the CIFAR100 training set.}
\label{tab:efficiency}
\centering
\resizebox{\linewidth}{!}{
\begin{tabular}{cccccc}
     \toprule
      \textbf{Sel. Ratio}& \textbf{K-Means} & \textbf{CoreSet} & \textbf{VAAL} & \textbf{LearnLoss}& \textbf{ours}\\ 
     \midrule
     $2\%$ & 16.6s& 1h57m& 7h52m & 20m & \textbf{12.6s}\\
     $5\%$ & 37.0s& 7h44m& 12h13m& 1h37m & \textbf{21.9s}\\
     $10\%$& 70.2s& 20h38m& 36h24m & 9h09m & \textbf{37.3s}\\
     \bottomrule
\end{tabular}
}
\end{table}

\subsection{Analysis}
\label{sec:analysis}
\textbf{Data Selection Efficiency} 
It is desirable that the data selection method operates in a time-efficient manner, as close as possible to random selection. In Tab.~\ref{tab:efficiency}, we compare the required time to select different percentages of training samples from CIFAR100. We do not take FDS into account due to its very poor performance. For those traditional active learning algorithms, both the repetitive model training and data sampling should be counted into the running time, and the former takes the majority. In contrast, our method chooses all the samples in a single-pass, so we do not have to train the model again in the selection process. As a result, our method's speed is faster than traditional active learning methods by a notable margin. Besides, unlike some active learning methods \cite{sener2017active,yoo2019learning}, our method does not need access to ground-truths before the end of all selection, which enables more flexibility of annotator assignment.


\begin{figure}[t!]
    \centering
    \begin{subfigure}{0.45\linewidth}
        \centering
        \includegraphics[width=\textwidth]{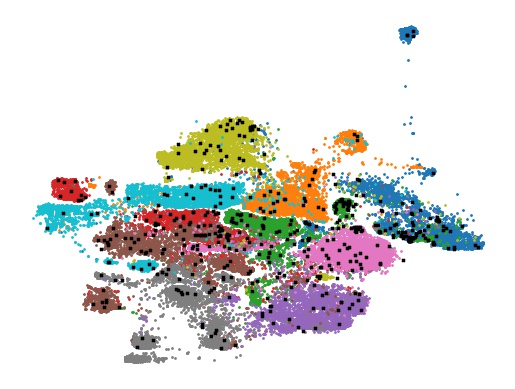}
        \caption{CoreSet}
    \end{subfigure}
    \begin{subfigure}{0.45\linewidth}
        \centering
        \includegraphics[width=\textwidth]{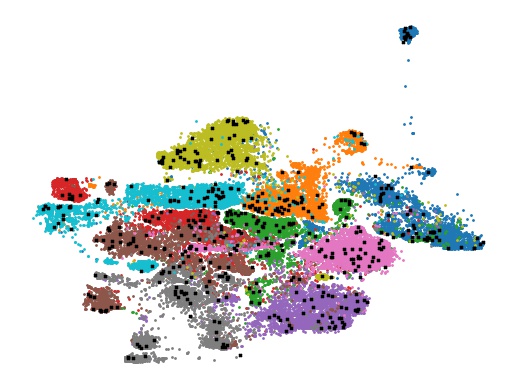}
        \caption{VAAL}
    \end{subfigure}
    \begin{subfigure}{0.45\linewidth}
        \centering
        \includegraphics[width=\textwidth]{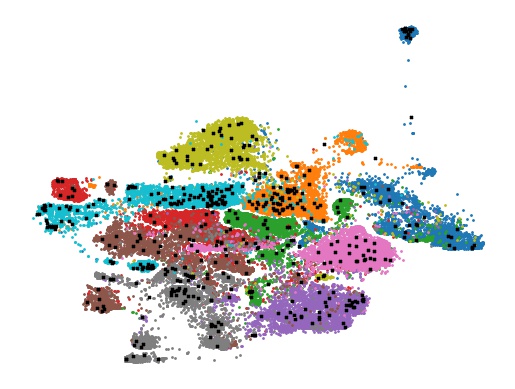}
        \caption{LearnLoss}
    \end{subfigure}
    \begin{subfigure}{0.45\linewidth}
        \centering
        \includegraphics[width=\textwidth]{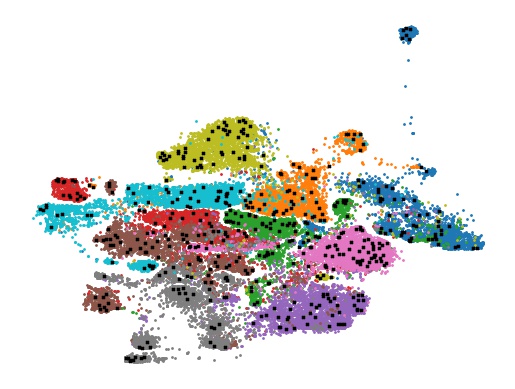}
        \caption{ActiveFT (ours)}
    \end{subfigure}
    \caption{\textbf{tSNE Embeddings of CIFAR10:} We visualize the embedding of selected samples using different algorithms. Different colors denote categories, and the black dots are the $1\%$ samples selected by our method.}
    \label{fig:visualization}
\end{figure}

\textbf{Visualization of Selected Samples} In Fig.~\ref{fig:visualization}, we visualize the feature $\mathbf{f}_i$ of each sample in CIFAR10 training set. The dimension of features is reduced by tSNE. The black dots represent the $1\%$ samples selected by different methods.
Results demonstrate that our selected samples distribute more similarly with the entire pool in the feature space than other counterparts. It reflects that optimizing our proposed parametric model helps to reduce the distribution gap between the selected subset and original unlabeled pool.

\textbf{Generality of our Method} ActiveFT can fit different pretraining frameworks and models well. In addition to DINO~\cite{caron2021emerging} framework and DeiT-Small~\cite{touvron2021training} model, we also apply ActiveFT to a DeiT-Small~\cite{touvron2021training} trained with generative unsupervised pretraining framework iBOT~\cite{zhou2021ibot} and CNN model ResNet-50~\cite{he2016deep} trained with DINO~\cite{caron2021emerging}. The models are pretrained on ImageNet-1k and would be finetuned in CIFAR10. Other implementation details are exactly same with Sec.~\ref{sec:classification}. Tab~\ref{tab:generality} shows the significant superiority of our results in comparison with random sampling baseline with
different sampling ratios. The results reflect the compatibility of ActiveFT with different unsupervised pretraining frameworks and model architectures.

\begin{table}[t!]
\caption{\textbf{Generality on Pretraining Frameworks and Model Architectures:} We examine the performance of ActiveFT on different pretraining frameworks and models on CIFAR-10.}
\label{tab:generality}
\centering
\begin{subtable}[t]{\linewidth}
    \caption{\textbf{Performance on DeiT-Small Pretrained with iBOT}}
    \label{tab:gen_deit}
    \centering
    \begin{tabular}{cccc}
        \toprule
        \textbf{Methods} & 0.5\% & 1\% & 2\%\\
         \midrule
         \textbf{Random} & 81.7 & 83.0 & 89.8\\
         \textbf{CoreSet~\cite{sener2017active}} & - & 82.8 & 89.2\\
        \textbf{LearnLoss~\cite{yoo2019learning}} & - & 83.6 & 89.2\\
        \textbf{VAAL~\cite{sinha2019variational}} & - & 85.1 & 89.3\\
        \midrule
        \textbf{ActiveFT (ours)} & \makecell[c]{\textbf{87.6}$\pm$0.8} &\makecell[c]{\textbf{88.3}$\pm$0.2} & \makecell[c]{\textbf{90.9}$\pm$0.2}\\
         \bottomrule
    \end{tabular}
\end{subtable}
\begin{subtable}[t]{\linewidth}
    \caption{\textbf{Performance on ResNet-50 Pretrained with DINO}}
    \label{tab:gen_resnet}
    \centering
    \begin{tabular}{cccc}
        \toprule
        \textbf{Methods} & 0.5\% & 1\% & 2\%\\
         \midrule
        \textbf{Random} & 64.8 & 76.2 & 83.7 \\
        \textbf{CoreSet~\cite{sener2017active}} & - & 70.4 & 83.2\\
        \textbf{LearnLoss~\cite{yoo2019learning}} & - & 71.7 & 81.3\\
        \textbf{VAAL~\cite{sinha2019variational}} & - & 75.0 & 83.3\\
         \midrule
        \textbf{ActiveFT (ours)} & \makecell[c]{\textbf{68.5} $\pm$0.4}& \makecell[c]{\textbf{78.6} $\pm$0.7}& \makecell[c]{\textbf{84.9} $\pm$0.3}\\
         \bottomrule
    \end{tabular}
\end{subtable}
\end{table}

\begin{table}[t!]
\caption{\textbf{Ablation Study:} We examine the effect of two modules in our method. Experiments are conducted on CIFAR100 with pretrained DeiT-Small model.}
\label{tab:ablation}
\centering
\begin{subtable}[t]{0.5\linewidth}
    \caption{\textbf{$c_i$ Update Manner}}
    \label{tab:ci}
    \centering
    \resizebox{\linewidth}{!}{
    \begin{tabular}{ccc}
        \toprule
        \textbf{Ratio} & \textbf{No-Update} & \textbf{Update}\\
        \midrule
        2\% & 20.6 & \textbf{40.7}\\
        5\% & 52.8 & \textbf{54.6}\\
        \bottomrule
    \end{tabular}
    }
\end{subtable}
\begin{subtable}[t]{0.46\linewidth}
    \caption{\textbf{Regularization Design}}
    \label{tab:negative}
    \centering
    \resizebox{\linewidth}{!}{
    \begin{tabular}{cccc}
        \toprule 
        \textbf{Ratio} & \textbf{S1} & \textbf{S2} & \textbf{ours}\\
        \midrule
        2\% & 33.1 & 26.8 & \textbf{40.7}\\
        5\% & 51.5 & 46.9 & \textbf{54.6}\\
        \bottomrule
    \end{tabular}
    }
\end{subtable}
\end{table}

\subsection{Ablation Study}
\label{sec:ablation}

\begin{table}[t!]
    \centering
    \caption{\textbf{Effect of Temperatures:} We try different temperatures in our method. Experiments are conducted on CIFAR10 with pretrained DeiT-Small model.}
    \label{tab:temperature}
    \begin{tabular}{cccccc}
        \toprule
         \textbf{Ratio} & $\tau=0.04$ & $\tau=0.07$ & $\tau=0.2$ & $\tau=0.5$\\
         \midrule
         0.5\% & \textbf{85.6} & 85.0 &84.1 & 83.5\\
         1\% & 87.4 & \textbf{88.2} & 85.3& 86.1\\
         2\% & \textbf{90.3} & 90.1 & 89.6 & 89.0\\
         \bottomrule
    \end{tabular}
\end{table}

We discuss the importance of different modules in our method including the update manner of $c_i$, the design of diversity regularization, and the effect of temperature.

\textbf{Update Manner of $c_i$} In this part, we discuss the ways to update $c_i$ (Eq.~\ref{eq:closest}) which denotes the parameter closest to each sample $\mathbf{f}_i$ in the feature space. In Alg.~\ref{alg:optimization}, it is updated in each iteration. Alternatively, we remain $c_i$ unchanged as the initial state in the optimization process. Results in Tab.~\ref{tab:ci} shows that this stationary strategy does not work well. In this case, it would rely heavily on the initial state. The frequent update of $c_i$ could help to relieve some harmful biases inside the initial state.

\textbf{Regularization Design} We try two alternative strategies to design the regularization term $R(\cdot)$ in Eq.~\ref{eq:loss}. \textbf{S1) No Regularization:} We only optimize the first term $D(\cdot,\cdot)$ in Eq.~\ref{eq:loss}.  \textbf{S2) InfoNCE~\cite{van2018representation}:} We get inspiration from \cite{van2018representation} to design a contrastive loss to approximate the distribution $p_{f_u}$ with $p_{\theta_{\mathcal{S}}}$:  $L=-\underset{\mathbf{f}_i\in\mathcal{F}^u}{E}\left[\log\frac{\exp(\mathbf{f}_i^T\theta_{\mathcal{S}}^{c_i}/\tau)}{\sum_{k\in[N]}\exp(sim(\mathbf{f}_k^T\theta_{\mathcal{S}}^{c_i}/\tau)}\right]$. In Tab.~\ref{tab:negative}, we evaluate these three strategies. We find that both \textbf{S1} and \textbf{S2} fails, and only our applied strategy \textbf{S3} succeeds. It justifies our design of the regularization strategy.

\textbf{Temperature $\tau$} We analyze the effect of different temperatures in Eq.~\ref{eq:loss}. Pointed out in Assumption~\ref{eq:large}, a small $\tau$ is a pre-requisite for our derivation. Tab.~\ref{tab:temperature} shows the results on CIFAR10 with different temperatures. When the temperature is relatively low (e.g. $\tau<$0.1), the performance of ActiveFT is great. However, as it becomes higher (e.g. $\tau=0.5$), the performance drops. The results are in line with our theoretical derivation.

\section{Conclusion} 
To fill in the gap inside the pretraining-finetuning paradigm, we define the active finetuning task, which selects samples from an unlabeled data pool for supervised model finetuning. To solve this problem, we propose a model-agnostic algorithm, ActiveFT. By optimizing a parametric model, ActiveFT chooses diverse data samples distributing similarly with the original pool for annotation. It is mathematically justified that ActiveFT helps to bring close the distributions of the selected subset and entire data pool by reducing the Earth Mover's distance. Our experiments on classification and segmentation show the state-of-the-art performance of ActiveFT, with an extremely high data selection efficiency. We believe ActiveFT can help to exploit the annotation budget for supervised finetuning in practical use and make a solid contribution to the popular pretraining-finetuning paradigms in various tasks.

{\small
\bibliographystyle{ieee_fullname}
\bibliography{arxiv}
}

\newpage

\begin{appendix}
In this appendix, we first analyze the effect of iteration number for the optimization of ActiveFT in Sec.~\ref{sec:ablation}. Then, we provide more implementation details in Sec.~\ref{sec:details}, including some explanation of N/A results in Tab.~\ref{tab:classification}. Finally, we give formal proof of the optimal joint distribution in Eq.~\ref{eq:transport} of our main paper in Sec.~\ref{sec:proof}.

\section{Ablation Study on Iteration Number}
We conduct an additional ablation study of the maximal iteration number $T$ (in Alg.~\ref{alg:optimization} of the main paper) of the parametric model optimization process in ActiveFT. The experiments are conducted on ImageNet \cite{russakovsky2015imagenet} with sampling ratio $1\%$. Results are demonstrated in Tab.~\ref{tab:iteration}. The quality of samples selected by ActiveFT continuously improves in the early stage as the optimization of our parametric model $p_{\theta_{\mathcal{S}}}$ goes, and then converges in the late stage. This result verifies that our model optimization gradually brings close the distributions of our selected samples to the entire unlabeled pool as well as ensures the diversity of the selected subset in the whole optimization process.

\begin{table}[htb]
\caption{\textbf{Ablation Study of Iteration Numbers:} Experiments are conducted on ImageNet \cite{russakovsky2015imagenet} dataset (1\% sampling ratio) with DeiT-Small \cite{touvron2021training} model pretrained with DINO \cite{caron2021emerging} framework. When iteration number is $0$, it is same as random selection.}
\label{tab:iteration}
\centering
\begin{tabular}{ccccccc}
     \toprule
     \multirow{2}{*}{\textbf{Sel. Ratio}}
     & \multicolumn{6}{c}{\textbf{Iteration Number}}\\
      & \textbf{0} & \textbf{50} & \textbf{75} & \textbf{100} & \textbf{200} & \textbf{300}\\
     \midrule
     1\% & 45.1 & 46.7 & 48.4 & \textbf{50.2} & 50.1 & 50.1 \\
     \bottomrule
\end{tabular}
\end{table}

\section{Additional Implementation Details}
\label{sec:details}
\subsection{Unsupervised Pretraining Details}
In our main paper, the DeiT-Small model (path size 16x16)  \cite{touvron2021training} is pretrained on ImageNet \cite{russakovsky2015imagenet} with DINO framework \footnote{https://github.com/facebookresearch/dino} \cite{caron2021emerging} for 300 epochs using AdamW optimizer \cite{loshchilov2018fixing} and batch size 1024. The learning rate is linearly ramped up to 5e-4$\times$batch\_size/256 in the first 10 epochs and decays with a cosine scheduler later. 

In Tab.~\ref{tab:generality} of our main paper, the DeiT-Small model \cite{touvron2021training} is pretrained with iBOT framework\footnote{https://github.com/bytedance/ibot} \cite{zhou2021ibot} on ImageNet \cite{russakovsky2015imagenet} for 800 epochs. The ResNet-50 model \cite{he2016deep} is pretrained with DINO framework \cite{caron2021emerging} on ImageNet for 300 epochs. The optimizer is AdamW \cite{loshchilov2018fixing} and the batch size is 1024 in both cases. The learning rate is linearly ramped up to 5e-4$\times$batch\_size/256 in the first 10 epochs too. 

\subsection{Supervised Finetuning Details}
\label{sec:finetune}
We typically follow the protocols in \cite{touvron2021training} to finetune the DeiT-Small model. For CIFAR10 and CIFAR100 \cite{krizhevsky2009learning} datasets, the pretrained models are supervisedly finetuned for 1000 epochs using SGD optimizer (lr=1e-3, weight-decay=1e-4, momentum=0.9) with batch size 512 and cosine learning rate decay on selected subsets of training data. For ImageNet \cite{russakovsky2015imagenet} dataset, to ensure convergence, the models are finetuned for 1000 epochs when the sampling ratio is 1\% and for 300 epochs when the sampling ratio is 5\%, using the same SGD optimizer as CIFAR. The images are resized to 224x224 in line with the pretraining. The supervised finetuning is implemented based on the official code of DeiT \footnote{https://github.com/facebookresearch/deit}. For ResNet-50 model in Tab.~\ref{tab:generality} of our main paper, we use the code base of mmclassification \footnote{https://github.com/open-mmlab/mmclassification}. We follow their settings to finetune the model with SGD optimizer (lr=1e-2, weight-decay=1e-4, momentum=0.9) with batch size 512 and cosine learning rate decay on selected subsets of training data for 100 epochs. 

On the semantic segmentation task, we follow \cite{strudel2021segmenter} to train the model for 127 epochs (\textit{i.e.} 16k and 32k iterations on 5\% and 10\% of training data). The model is trained using SGD optimizer (lr=1e-3, momentum=0.9) with batch size 8 and polynomial learning rate decay. The code base is mmsegmentation \footnote{https://github.com/open-mmlab/mmsegmentation}.

\subsection{Active Learning Transplantation Details}
\label{sec:active}
We transplant three classical active learning methods and two newer algorithms to the pretraining-finetuning paradigm, including CoreSet \cite{sener2017active}, VAAL \cite{sinha2019variational}, LearnLoss \cite{yoo2019learning}, TA-VAAL \cite{kim2021task}, and ALFA-Mix \cite{parvaneh2022active}.

For all five methods, we apply them to image classification task on CIFAR10, CIFAR100 and ImageNet. These methods select data samples with batch-selection strategy. Firstly, we train the model on a randomly sampled initial set. Then, the model is used to select a batch of images from the training set, and the model is re-trained on all the selected samples. This process repeats until the annotation budget is filled. In the pretraining-finetuning paradigm, for CoreSet, LearnLoss and ALFA-Mix, we use DeiT-Small \cite{touvron2021training} pretrained with DINO \cite{caron2021emerging} as the backbone of their models for data selection. For VAAL and TA-VAAL, we directly use their original light-weighted VAE to select data. When the data samples have been selected with different sampling ratios, we finetune the DeiT-Small model in the same manner as Sec.~\ref{sec:finetune} on the selected data samples. The sizes of the initial set and each selection batch are set as 0.5\% on CIFAR10, 1\% on CIFAR100, and 2.5\% on ImageNet separately for all the five algorithms.

\subsection{Explanation of N/A Results}
There are some N/A results (denoted by ``-") in Tab.~\ref{tab:classification} of our main paper. We explain them from the following three angles.
\begin{itemize}
    \item \textbf{Initial Set of Active Learning:} Described in Sec.~\ref{sec:active}, all five active learning methods require to randomly sample a small initial set in the beginning. On this initial set, the performance of these active learning algorithms is same as random sampling. Therefore, we pass the duplicate results on these random initial sets \textit{i.e.} 0.5\% of CIFAR10 and 1\% of CIFAR100. Since $1\%$ is smaller than the initial set size (2.5\%) on ImageNet, we pass this sampling ratio as well.
    \item \textbf{K-Means on ImageNet:} Given the large number of images in training set, it is hard to implement K-Means to ImageNet dataset, which exceeds the capability of our hardware. Since K-Means does not perform well on CIFAR10 and CIFAR100, the N/A results on ImageNet would not affect our conclusions.
\end{itemize}

\section{Proof of the Optimal Distributions for Earth Mover’s Distance}
\label{sec:proof}
In Sec.~\ref{sec:math} of our main paper, we give an optimal distribution to calculate the earth mover’s distance (EMD), \textit{i.e.} each $\mathbf{f}_i\sim p_{f_u}$ transports to their closest $\mathbf{f}_{s_j}\sim p_{f_{\mathcal{S}}}$. Eq.~\ref{eq:transport} in the main paper is copied as follows:
\begin{equation}
    \gamma_{f_u,f_S}(\mathbf{f}_i,\mathbf{f}_{s_j})=\begin{cases}
    \frac{1}{N} & \mathbf{f}_i\in \mathcal{F}^u,\mathbf{f}_{s_j}\in\mathcal{F}^u_{\mathcal{S}},c_i=j\\
    0 & otherwise\\
    \end{cases}
    \label{eq:joint}
\end{equation}

We will prove it is the optimal joint distribution $\gamma$ to reach the infimum in Eq.~\ref{eq:wdist} of our main paper, copied as follows:
\begin{equation}
    EMD(p_{f_u},p_{f_S})=\underset{\gamma\in\Pi(p_{f_u},p_{f_S})}{inf}\underset{(\mathbf{f}_i,\mathbf{f}_{s_j})\sim \gamma}{E}\left[||\mathbf{f}_i-\mathbf{f}_{s_j}||_2\right]
\end{equation}
Suppose there is a general format:
\begin{equation}
    \gamma_{f_u,f_S}(\mathbf{f}_i,\mathbf{f}_{s_j})=
    p(\mathbf{f}_i,\mathbf{f}_{s_j}) \qquad \mathbf{f}_i\in \mathcal{F}^u,\mathbf{f}_{s_j}\in\mathcal{F}^u_{\mathcal{S}}\\
\end{equation}

Because of the uniform distribution of $p_{f_u}$, $p(\mathbf{f}_i,\mathbf{f}_{s_j})$ satisfies the following constraints.
\begin{equation}
    p(\mathbf{f}_i,\mathbf{f}_{s_j}) \geq 0, \qquad \sum_{\mathbf{f}_{s_j}\in\mathcal{F}^u_{\mathcal{S}}} p(\mathbf{f}_i,\mathbf{f}_{s_j})=p_{f_u}(\mathbf{f}_i) = \frac{1}{N}.
\end{equation}

The distance expectation for each feature $\mathbf{f}_i$ with the distribution $\mathcal{F}^u$ is 
\begin{equation}
    \begin{aligned}
        \underset{\mathbf{f}_{s_j}\in\mathcal{F}^u_{\mathcal{S}}}{E} \left[||\mathbf{f}_i-\mathbf{f}_{s_j}||_2\right] =& \sum_{\mathbf{f}_{s_j}\in\mathcal{F}^u_{\mathcal{S}}} \left[p(\mathbf{f}_{s_j}|\mathbf{f}_i) \cdot ||\mathbf{f}_i-\mathbf{f}_{s_j}||_2 \right]\\
        =&\sum_{\mathbf{f}_{s_j}\in\mathcal{F}^u_{\mathcal{S}}} \left[p(\mathbf{f}_i, \mathbf{f}_{s_j})/p_{f_u}(\mathbf{f}_i) \cdot ||\mathbf{f}_i-\mathbf{f}_{s_j}||_2 \right]\\
        =&N\sum_{\mathbf{f}_{s_j}\in\mathcal{F}^u_{\mathcal{S}}} \left[p(\mathbf{f}_i, \mathbf{f}_{s_j}) \cdot ||\mathbf{f}_i-\mathbf{f}_{s_j}||_2 \right]
    \end{aligned}
\end{equation}

Since the $\mathbf{f}_{s_{c_i}}$ is the nearest feature of $\mathbf{f}_i$ in $\mathbf{f}_{s_j}\sim\mathcal{F}^u_{\mathcal{S}}$, we can have the inequality
\begin{equation}
    ||\mathbf{f}_i-\mathbf{f}_{s_j}||_2\geq||\mathbf{f}_i-\mathbf{f}_{s_{c_i}}||_2
\end{equation}
so
\begin{equation}
\begin{aligned}
    \underset{\mathbf{f}_{s_j}\in\mathcal{F}^u_{\mathcal{S}}}{E} \left[||\mathbf{f}_i-\mathbf{f}_{s_j}||_2\right] \geq& N\cdot \left[ \sum_{\mathbf{f}_{s_j}\in\mathcal{F}^u_{\mathcal{S}}} p(\mathbf{f}_i,\mathbf{f}_{s_j}) \right] \cdot ||\mathbf{f}_i-\mathbf{f}_{s_{c_i}}||_2\\
    =&N\cdot\frac{1}{N}||\mathbf{f}_i-\mathbf{f}_{s_{c_i}}||_2\\
    =&||\mathbf{f}_i-\mathbf{f}_{s_{c_i}}||_2
\end{aligned}
\end{equation}
The above minimum is reached when Eq.~\ref{eq:joint} (Eq.~\ref{eq:transport} in our main paper) is satisfied. Therefore, for each $\mathbf{f}_i$, we only need to find the nearest feature $\mathbf{f}_{s_{c_i}}$ among $\mathbf{f}_{s_j}\sim\mathcal{F}^u_{\mathcal{S}}$ and assign the joint probability as $\frac{1}{N}$.

\end{appendix}

\end{document}